\documentclass[runningheads]{llncs}
\usepackage{graphicx}
\usepackage{comment}
\usepackage{enumitem}
\usepackage{bigstrut}
\usepackage{xcolor}
\begin{document}
\title{ I Can't Believe It's Not Better: In-air Movement For Alzheimer Handwriting Synthetic Generation}

%
%

\author {Asma Bensalah\inst{1,2}\orcidID{0000-0002-2405-9811} \and
Antonio Parziale\inst{3,4}\orcidID{0000-0003-0858-3397} \and
Giuseppe De Gregorio\inst{3,4}\orcidID{0000-0002-8195-4118} \and
Angelo Marcelli\inst{3,4}\orcidID{0000-0002-2019-2826} \and
Alicia Fornés\inst{1,2}\orcidID{0000-0002-9692-5336} \and
Josep Lladós\inst{1,2}\orcidID{0000-0002-4533-4739}}
\authorrunning{A. Bensalah et al.}
%
\institute{Computer Vision Center, Universitat Autònoma de Barcelona, Spain \and
\email{\{afornes, abensalah, josep\}@cvc.uab.es}\\
Computer Science Department, Universitat Autònoma de Barcelona, Spain \and
DIEM, University of Salerno, Via Giovanni Paolo II 132, 84084 Fisciano, SA, Italy\\
\email{\{anparziale, gdegregorio , amarcelli\}@unisa.it} \and
AI3S Unit, CINI National Laboratory of Artificial Intelligence and Intelligent
Systems, University of Salerno, Fisciano, SA, Italy}

\maketitle              
\begin{abstract}
During recent years, there here has been a boom in terms of deep learning use for handwriting analysis and recognition. One main application for handwriting analysis is early detection and diagnosis in the health field. Unfortunately, most real case problems still suffer a scarcity of data, which makes difficult the use of deep learning-based models. To alleviate this problem, some works resort to synthetic data generation. Lately, more works are directed towards guided data synthetic generation, a generation that uses the domain and data knowledge to generate realistic data that can be useful to train deep learning models. In this work, we combine the domain knowledge about the Alzheimer's disease for handwriting and use it for a more guided data generation. Concretely, we have explored the use of in-air movements for synthetic data generation.
\keywords{In-air Movements \and online handwriting recognition \and Synthetic Data Generation \and Alzheimer disease \and Recurrent Neural Networks \and Convolutional Neural Networks.}
\end{abstract}
\section{Introduction}
Deep learning models are data hungry. Hence, in many application scenarios, additionally to collecting more data specific to the relevant task, generating samples synthetically has been widely adopted as an alternative to alleviate the few data issue~\cite{synthetic}. It is well known that when it comes to tasks related to the medical field, data scarcity is more severe since collecting data in clinical setups is tough and there exist privacy preservation concerns~\cite{health_data}. 

Alzheimer's disease (AD) can be defined as a slowly progressive irreversible degenerative disease with well-defined pathophysiological mechanisms~\cite{ad1}. AD is marked by a decrease in cognitive skills and the individual's independence levels when performing daily life activities~\cite{ad}; besides being the most common reason behind dementia.

Early AD detection is essential for screening purposes and later AD patients' disease management. Moreover, it serves to help the patients and their caregivers to plan for the future thus help the patient to maintain a desired quality of life as long as possible. Very often, AD diagnosis in clinical practice can be complicated due to time constraints and due to the fact that AD symptoms can be considered as normal aging symptoms. Furthermore, AD early diagnosis can lessen the financial cost related to AD patients' and their caregivers' support ~\cite{ad2}~\cite{ad3}. 
In this scenario, handwriting analysis remains an affordable and efficient alternative for AD early diagnosis and detection contrary to other AD early diagnosis and detection approaches such as invasive and non-invasive biomarkers methods. The latter are generally expensive and limited in terms of availability in clinical practice. In addition, there is a need for special 
expertise when dealing with technologies that perform invasive biomarkers examinations~\cite{can}. 

Furthermore, it is well known that handwriting problems arise among neurodegenerative patients and AD patients in particular~\cite{hand_ad}. For instance, small handwriting size referred to as micrographia is linked to Parkinson's disease (PD), meanwhile dysgraphia, defined as the neurological condition that cripples writing abilities~\cite{dysgraphia}, is observed among AD patients~\cite{dynamic,ref2,ref3,ref4}. Hence, handwriting could be deemed an important biomarker to diagnose AD~\cite{diagnosis,diagnosis1}.
It is assumed that smoother velocity profiles mean more efficient neuromotor systems. Indeed, upon this assumption was built the poor handwriting theory~\cite{galen}. The theory claims that once a motor system fails to limit the noise behind the accelerate and decelerative forces, it is not kinetically optimal and unpredictable spatially. 
 
Werner et al~\cite{in-air} examined kinematically the handwriting process amid mild cognitive impairment (MCI), mild AD, and healthy populations. In addition, they assessed the relative significance of handwriting's kinematic features across the three populations. The authors found out that, apart from velocity, all kinematic measures consistently differentiate between healthy and AD individuals.
 
One interesting finding shown in~\cite{in-air} was the increase in in-air time within AD and MCI groups compared to the control group (healthy). In-air time was defined as the time when the pen is not on whatever writing support used (e.g. paper, tablet, etc.). Several reasons were cited to interpret this outcome: the first one was related to the writing characteristics of the language used for the experiments (concretely, Hebrew) which already requires more pen lifts than Latin-based languages due to the specific language writing characteristics. The second one was related to the theoretical model of Van Galen and Teulings~\cite{van+teulings}, in which the model discerns three phases in the writing response: motor programming (patterns retrieval), parameter setting, and motor initiations (impulse generation for particular muscles). On the ground of these three steps, it could be deduced that the increase in in-air time is due to the deficit in motor programming amid AD patients who take relatively a longer time to start a movement~\cite{response}~\cite{response1}. Finally, the authors suggested that visuospatial deficit among AD patients could be a possible reason~\cite{visuospatial,visuospatial1}.
 
\begin{figure}[t]
  \centering
  \begin{tabular}{cc}
    \includegraphics[width=0.5\textwidth, height=4cm]{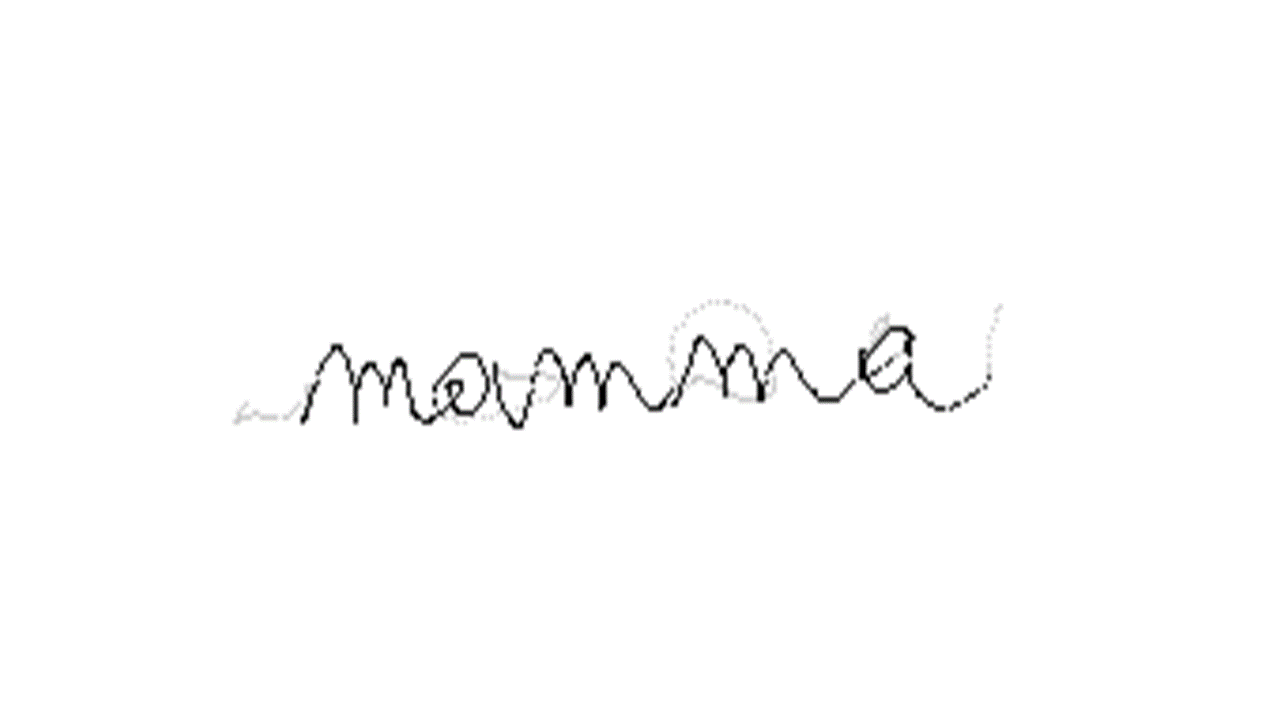} &
    \includegraphics[width=0.5\textwidth, height=4cm]{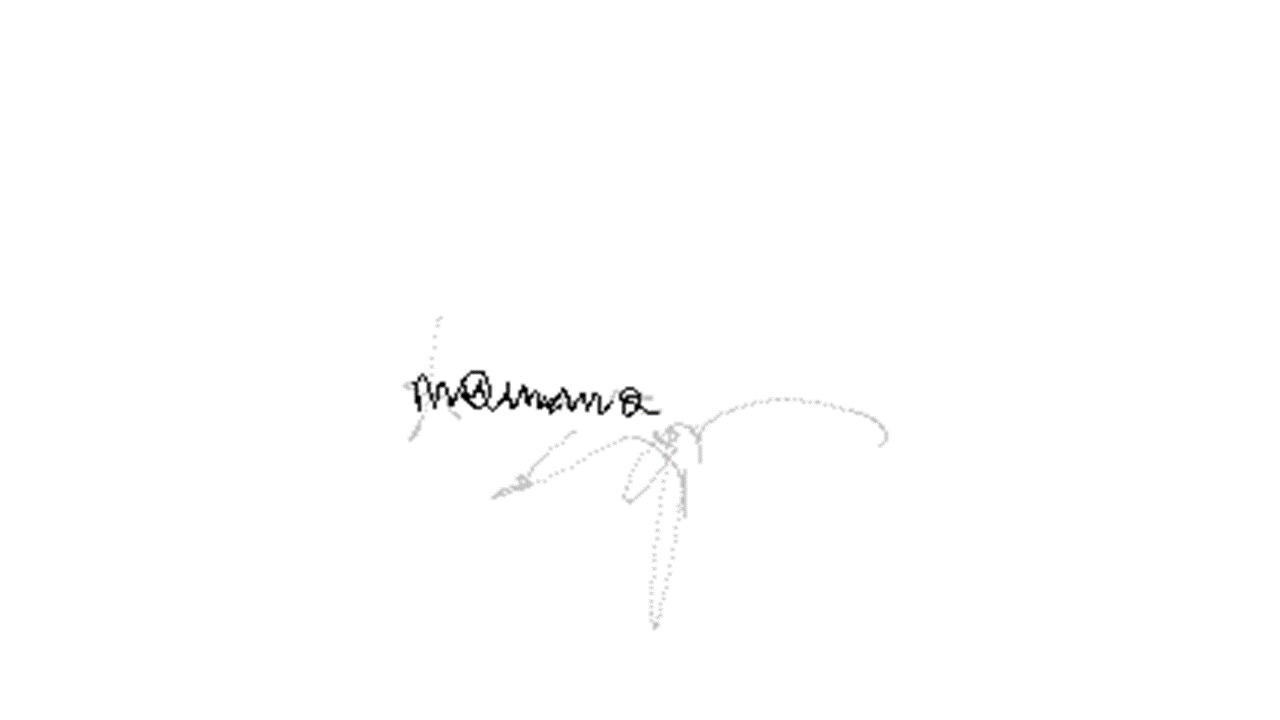} \\
  a) & b)  \\
  \end{tabular}\\
  \caption{On-paper (black) and in-air movements (gray) for the word "Mamma" written above a line. a) Healthy individual 1; b) AD patient 2. }
    \label{fig:mamma}
\end{figure}

\begin{figure}[t]
  \centering
  \begin{tabular}{cc}
    \includegraphics[width=0.5\textwidth, height=4cm]{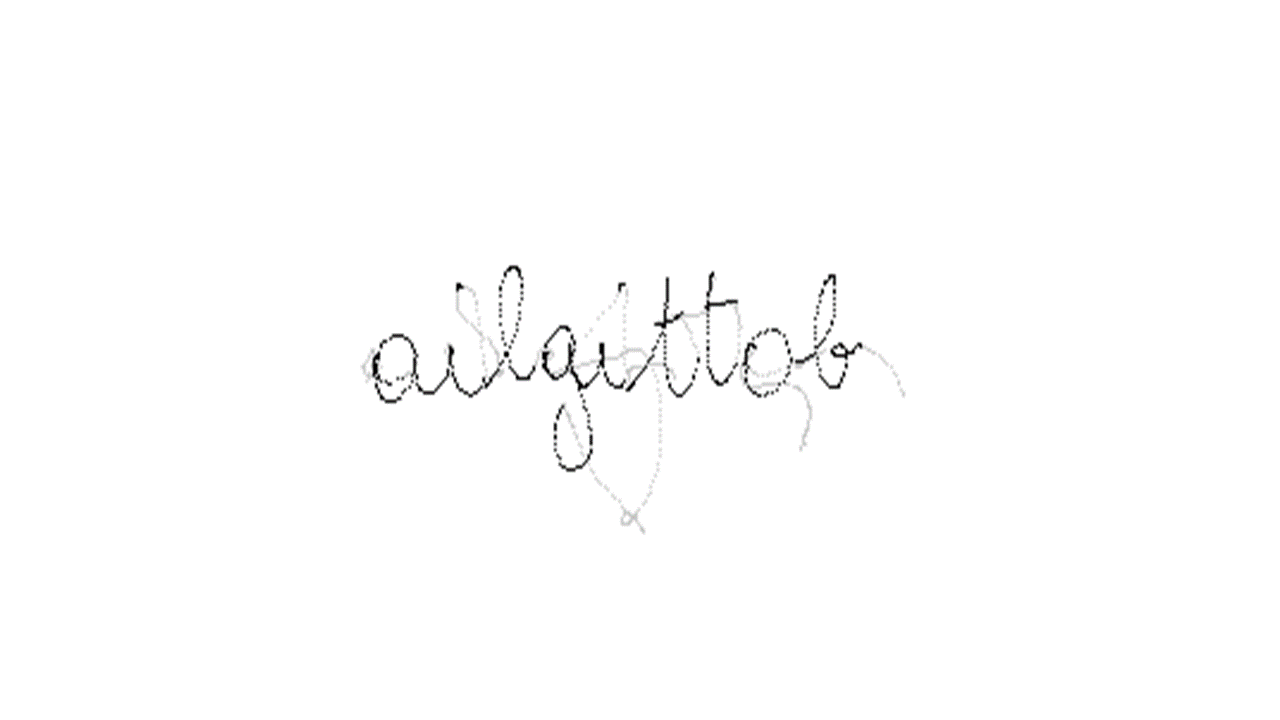} &
    \includegraphics[width=0.5\textwidth, height=4cm]{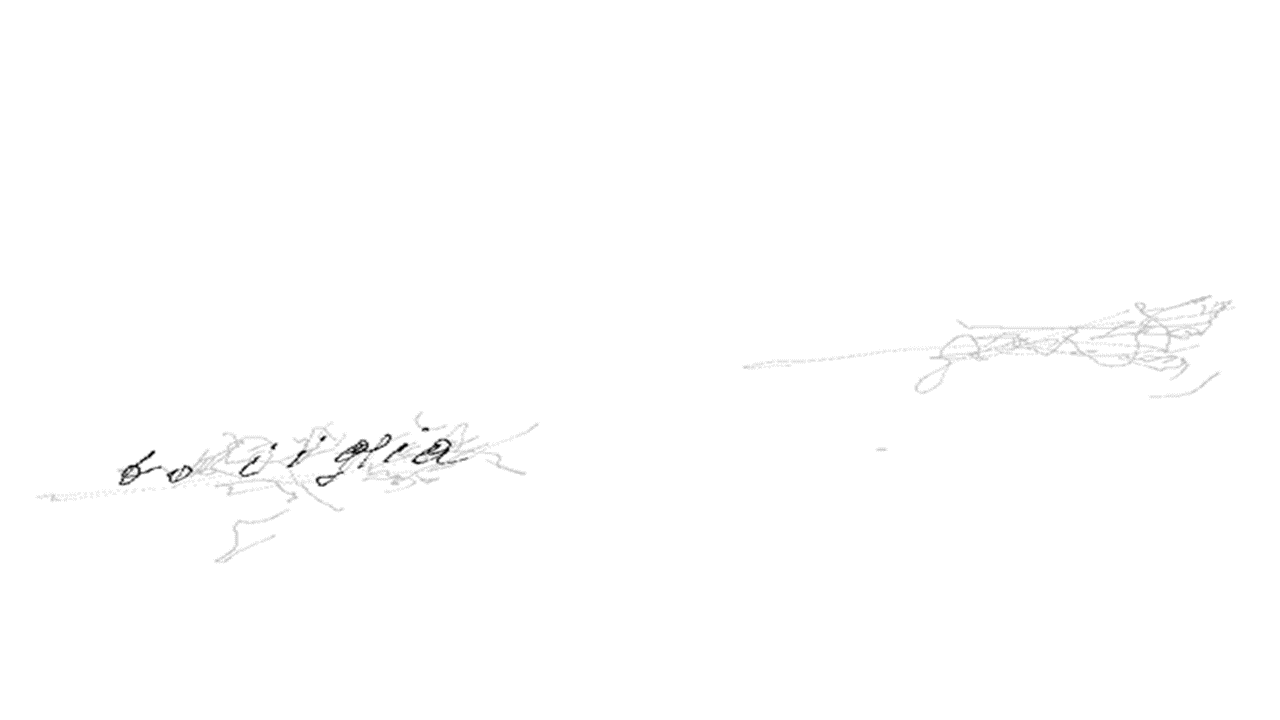} \\
  a) & b)  \\
  \end{tabular}\\
  \caption{On-paper (black) and in-air movements(gray) for the word "bottiglia" written in reverse . a) Healthy individual 1; b) AD patient 2. }
    \label{fig:bottle}
\end{figure}

Therefore, we are impelled by this work on the relevant discrimination power of in-air movements in AD patients' handwriting and our observations across many datasets for neurodegenerative diseases (see Figures~\ref{fig:bottle} and \ref{fig:mamma}) into the bargain, coupled with synthetic generation to face the data scarcity for neurodegenerative diseases. In essence, we explore in this work the use of in-air movements for synthetic handwriting generation. Our initial hypothesis is that in-air movement information could lead to generating good-quality synthetic samples and therefore, models trained with this data can reach better classification accuracies.
 
To our best knowledge, handwriting synthetic generation for AD detection and diagnosis, in particular, is not a prosperous research area, which requires more research efforts from the scientific community. 
 
The rest of the article is organized as follows: Section~\ref{sec:related} reviews relevant related works. Next, Section~\ref{sec:methodology} describes the generator/discriminator duality in our implementation. Afterwards, experiment details are given in Section~\ref{sec:experiments} for reproducibility purposes. Then, Section~\ref{sec:results} presents the different results, which are discussed later in Section~\ref{discussion}. Finally, we draw conclusions in Section~\ref{conclusion}.

\section{Related work}~\label{sec:related}
Many works covered online handwriting analysis for neurodegenerative patients as it offers more information about the individual's kinematics and fine motor skills than offline analysis. 
Most of those works take into account kinematic features to automatically discriminate between a control group of healthy individuals and a patient group~\cite{dynamic,hand_parkinson}. All those works fall into the line of Computer Aided Diagnosis (CAD) systems which help clinicians and doctors by providing biomarker computations and analysis. CAD systems could be integrated in already existing clinical workflows in order to maximize early diagnosis and detection chances~\cite{cad}.

The analyzed tasks can be classified into the following categories:
\begin{itemize} [label=$\bullet$]
    \item Drawing tasks: individuals are asked to draw different forms: spirals~\cite{spiral}, meanders, lines~\cite{lines}, etc.
     \item Writing tasks: individuals are asked to write a letter or a sequence of letters, words, sentences~\cite{dataset}, etc.
     \item Complex tasks: individuals are asked to perform the writing/drawing task in addition to another motor/cognitive task with the purpose of increasing the task load, thus revealing more motor/cognitive issues~\cite{complex}.  
\end{itemize}

Traditionally, statistical tests (for instance, ANOVA) are used to analyze online handwriting~\cite{anova1,anova2}. In the past years, the handwriting analysis area has benefited from the machine/deep learning boom, in particular handwriting analysis for CAD systems: neurodegenerative diseases and others~\cite{deep1,deep2}.

Some works have tackled synthetic sample handwriting generation for neurodegenerative diseases~\cite{generation1,igs}.

Nevertheless, there is still a gap in data generation for Alzheimer's disease, in particular, and a lack of works that focus on the domain knowledge for guided data generation.

\section{Methodology}~\label{sec:methodology}
As explained in the introduction, we aim to evaluate the impact of in-air movements in the generation of synthetic samples and in the classification tasks. 

We have used the architecture presented in~\cite{igs} to generate and select the synthetic samples used for training the classifier that discriminates between the handwriting sample drawn by AD patients or healthy subjects.
The architecture consists of a Generator (RNN), trained to generate new samples, and a Discriminator (CNN network), which classifies the generated images into fake or real samples. Once the discriminator can be fooled, the generation is supposed to generate good-quality data that can be used with real data to train the final classifier.

Contrary to conventional GANs architectures, the GAN loss does not back-propagate through the generator to update each layer's weights. Instead, the Generator and Discriminator in this architecture are parallel. Thus the generator does not learn from the discriminator's feedback (code compatibility reasons).

The two modules are further described in the next subsections.



\subsection{Generator}
First, the real data is organised into 5 folds, and used later for training and testing. Then, synthetic images are generated.
The generator is inspired by Alex Graves's work~\cite{Rnn}, where recurrent neural networks (RNNs) were used to generate realistic handwriting sequences. Rather than having the RNNs model predict exactly what the future point will be, Graves's work discusses predicting a probability distribution of the future given the prior information.

The generator is fed with two channels of the input time series: acceleration through x ($a_{x}$) and y ($a_{y}$) axis. A two-layer stacked basic LSTM has been used, with 256 nodes in each layer.
The generator output is a sequence of SL points, where SL is chosen so that the distribution of synthetic samples per number of points was similar to the distribution of real samples per number of points.

\subsection{Discriminator}
The discriminator validates the data outputted by the generator. A synthetic image is hold when the discriminator assigns to it the correct class, otherwise is discarded.

The discriminator is an ensemble of 5 Convolutional Neural Networks (CNNs) and it classifies samples by a majority vote rule. The dataset of real samples is shuffled 5 times and each time one of the CNN belonging to the ensemble is trained with 35\% of the data.

\begin{figure}[ht!]
\centering
\includegraphics[ height=0.9\textheight]
{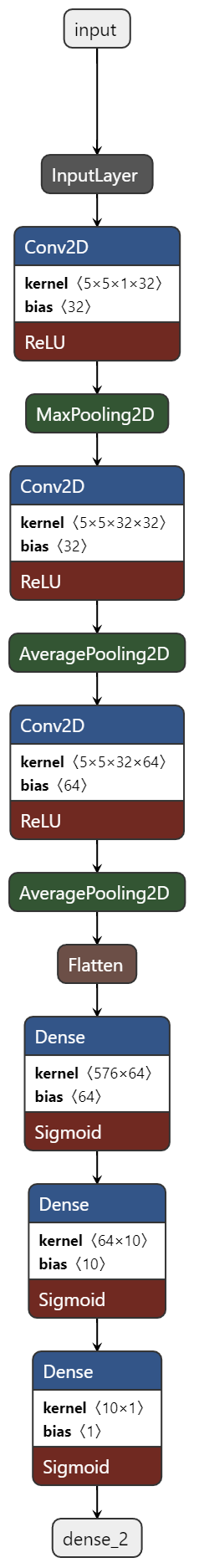}
\caption{One of the 5 CNN that make up the Discriminator architecture}
\label{fig:discriminator}
\end{figure}

Each CNN of the ensemble is made-up of 5 convolutional layers and its architecture is equal to the CIFAR-10 neural network presented in \cite{Pereira:CMPB16}. Figure~\ref{fig:discriminator} shows the neural network and the hyper-parameters related to each layer.
Table~\ref{tab:classification_parameters} reports the hyper-parameter values chosen for training the 5 CNNs.

The adopted discriminator elaborates 2D images so, both the real and generated time series are converted into 2D grayscale images, as described in \cite{igs}. In particular, the time series of each real or synthetic handwriting sample are rearranged into a squared matrix that is then resized in a 64 $\times$ 64 image using the Lanczos re-sampling method.


\section{Dataset}~\label{sec:dataset}

The DARWIN dataset was introduced in~\cite{dataset} as the largest publicly available dataset~\cite{website} in terms of participants' numbers. The dataset includes handwriting samples from 174 individuals: 89 AD patients and 85 healthy individuals. To ensure maximum pattern matching between the two groups, participants from both groups have the same age distribution, educational background, and work profession type (intellectual/manual). 

The dataset was obtained by writing with a pen on an A4 sheet of white paper placed over a Wacom's Bamboo tablet. The recorded information is: 
\begin{enumerate}
    \item timestamp;
    \item x coordinate;
    \item y coordinate;
    \item binary pen-down property;
    \item the pressure applied by the pen on the paper.
\end{enumerate}

Individuals were asked to perform 25 tasks, for instance:
\begin{itemize}
    \item Joining two points with a horizontal line, continuously four times; 
\item Retracing a circle (6 cm in diameter) continuously four times; 
\item Copying the word ‘‘foglio’’; 
\item Copying the letters ‘l’, ‘m’, and ‘p’; 
\item Copying the letters on the adjacent rows; 
\item Writing cursively a sequence of four lowercase letters ‘l’, in a single smooth movement; 
\item Writing cursively a sequence of four lowercase cursive bigram ‘le’, in a single smooth movement; 
\item Copying in reverse the word ‘‘casa’’; 
\item Drawing a clock with all hours and putting hands at 11:05 (Clock Drawing Test); 
\item Copying a paragraph;
\end{itemize}

For our work, we are interested in particular by two of these tasks:
\begin{itemize} [label=$\bullet$]
    \item Task 13: Copy the word ‘mamma' (the Italian word for mom) above a line;
    \item Task 16: Copy in reverse the word ‘bottiglia’ (the Italian word for bottle);
\end{itemize}

Task 13 has been chosen over many works dealing with neurodegenerative diseases handwriting analysis~\cite{mamma} because of its significant presence in someone's language since early childhood, besides the fact that it is a word commonly repeated by AD patients in advanced disease stages. 

On the other hand Task 16 is an interesting task because it is a complex one since it consists of a word reverse copying which implies a cognitive effort (inspired from the Mini Mental State Examination).

\section{Experiments}~\label{sec:experiments}
We define two experimentation scenarios for each Task:
\begin{itemize}
    \item In-air movements: using only movements performed with the pen when it is not on writing support;
    \item In-air + On-paper: using both movements recorded when the pen is and is not on the writing support;
\end{itemize}
Initially, authors in~\cite{igs} have found that feeding the generator with more than two channels from the input time series has weak effects on the method's performance. For the same reason and for optimal memory and computation time, we have chosen the $a_{x}$ and $a_{y}$ channels. For reproducibility uses, Table~\ref{tab:hyper} describes the generator's hyperparameters.

It's worth nothing that for each scenario and for each task we trained 2 different RNNs: the first synthesized samples drawn by healthy subjects, the second synthesized samples drawn by AD patients.

Moreover, for each scenario and for each task, a CNN that discriminates between samples drawn by a healthy subject or AD patient was trained with the hyperparameters reported in Table~\ref{tab:classification_parameters} and using both real and synthetic samples. The number of generated synthetic samples has been either 500 (500 AD and 500 healthy) or 1000 (1000 AD and 1000 healthy), for each task. 
The performance was measured by averaging on 5 training of the CNN. At each training, the real dataset was shuffled and 50\% of subjects were kept apart as test set.

\begin{table}[htbp]
  \centering
  \caption{Generator's hyperparameters.}
    \begin{tabular}{ll}
    \textbf{Parameter} & \textbf{Chosen Value} \bigstrut[b]\\
    \hline
    RNN hidden state & 256 \bigstrut[t]\\
    Number of layers & 2 \\
    Cell Type & LSTM \\
    SL  & 150 \\
    Number of epochs & 301 \\
    Learning rate & 0.01 \\
    Number of Mixture M & 20 \\
    Dropout keep probability & 0.8 \\
    Training/validation set & (70\%,30\%)  \\
    Loss Function & Log likelihood loss \bigstrut[b]\\
    \hline
    \end{tabular}%
  \label{tab:hyper}%
\end{table}%

\begin{table}[tb!]
\centering
\caption{Experimental setup to classify 2D images with the CNN}
\label{tab:classification_parameters}
\begin{tabular}{|l|c|} 
\hline
\textbf{Parameter}  & \textbf{Value} \\
\hline
\textbf{Kernel Initializer}  & Glorot Normal \\
\textbf{Bias Initializer} &  0 \\
\textbf{\textbf{Pseudorandom number generators}}& Fixed Seeds \\
\textbf{Training / Validation} & 35\% / 15\%\\
\textbf{k-fold cross validation} &  5-fold \\
\textbf{\textbf{Batch size}} &  5 \\
\textbf{Optimization~algorithm} &  SGD \\
\textbf{Learning Rate} &  $2\times10^{-5}$ \\
\textbf{Momentum} &  0.9 \\
\textbf{Nesterov Momentum} &  True \\
\textbf{Loss} &  Binary Cross Entropy \\
\textbf{Early stopping} &  Min Validation Loss \\
\textbf{Epochs} &  10000 \\
\hline
\end{tabular}
\end{table} 
\section{Results}~\label{sec:results}
Table~\ref{tab:mama_acc_500} provides the average classification accuracies for both Task 13 (mamma) and Task 16 (bottiglia) when generating 500 synthetic samples, while Table~\ref{tab:bottle_acc_1000} compares the average accuracies when generating 1000 synthetic samples. In both cases, we compare the performance of there scenarios using: in-air movements, on-paper movements and in-air movements together with on-paper movements.

\begin{table}[htbp]
  \centering
  \caption{Average accuracies for Task 13 and Task 16 using 500 synthetic samples.}
    \begin{tabular}{crrrrr}
          & \multicolumn{5}{c}{500  synthetic samples} \bigstrut[b]\\
\cline{2-6}          & In-air &       & on-paper &       & In-air+on-paper \bigstrut\\
\cline{2-6}    Task13  (mamma)  & 35,71\% &       & 43,77\% &       & 45,15\% \bigstrut[t]\\
    Task16 (bottiglia) & 45,15\% &       & 54,66\% &       & 51,46\% \\
    \end{tabular}%
  \label{tab:mama_acc_500}%
\end{table}%

\begin{table}[htbp]
  \centering
  \caption{Average accuracies for Task 16 using 1000 synthetic samples.}
    \begin{tabular}{rrrrrr}
          & \multicolumn{5}{c}{1000  synthetic samples} \bigstrut[b]\\
\cline{2-6}          & In-air &       & On-paper &       & In-air+On-paper \bigstrut\\
\cline{2-6}    Task16 (bottiglia) & 51,59\% &       & 54,14\% &       & 52,02\% \bigstrut[t]\\
    \end{tabular}%
  \label{tab:bottle_acc_1000}%
\end{table}%

Although we observe that there's a significant decrease in terms of average accuracy for Task 13 when using in-air movements only to generate synthetic samples, Table~\ref{tab:bottle_acc_1000} shows that accuracies remain almost the same when using in-air and in-air+ on-paper movements for sample generation with a tiny difference of 0,43\%.
We observe that using in-air movements, the average classification accuracy is higher for Task 16 compared to Task 13 (Table~\ref{tab:mama_acc_500}).

It is interesting to notice that while for task 13 using in-air and on-paper movements together leads to better accuracies (45,15\%) compared to when they are used separately (35,71\%, 43,77\%) to generate synthetic data , it is not the case for task 16. This pattern is similar for the case of task 16, when more synthetic samples are generated (see Table~\ref{tab:bottle_acc_1000}), the best accuracy is still achieved with on-paper movements solely.

For comparison purposes, Table~\ref{tab:no_synthetic} provides the average accuracies when using in-air movements per fold when no synthetic samples are generated. It can be observed that the average accuracy reaches 56,78\% for Task 13 while it is higher by 0,34\% for Task 16.

\begin{table}[htbp]
  \centering
  \caption{Average accuracies using in-air movements per Fold with no synthetic data.}
    \begin{tabular}{ccccccc}
    \multicolumn{7}{c}{\textbf{In-air (No synthetic samples)}} \bigstrut[b]\\
    \hline
    \textbf{Accuraccy } & \textbf{Fold 1} & \textbf{Fold 2} & \textbf{Fold 3} & \textbf{Fold 4} & \textbf{Fold 5} & \textbf{Average} \bigstrut\\
    \hline
    Task 13 (mamma)  & 62,50\% & 57,14\% & 50\%  & 50\%  & 64,28\% & \textbf{56,78\%} \bigstrut[t]\\
    Task 16 (bottiglia) & 50\%  & 71,42\% & 52,94 & 68,42\% & 42,85 & \textbf{57,12\%} \\
    \end{tabular}%
  \label{tab:no_synthetic}%
\end{table}%

\section{Discussion}~\label{discussion}
Our initial hypothesis was that in-air movements represent discriminative patterns for the Alzheimer's Disease patient population. Overall, the results show that there is a gap between classification accuracies when using only in-air movements versus the use of in-air plus on-paper movements and this gap depends on the task complexity and the number of synthetic samples.
 
\begin{figure}[t]
  \centering
  \begin{tabular}{cc}
    \includegraphics[width=0.5\textwidth, height=4cm]{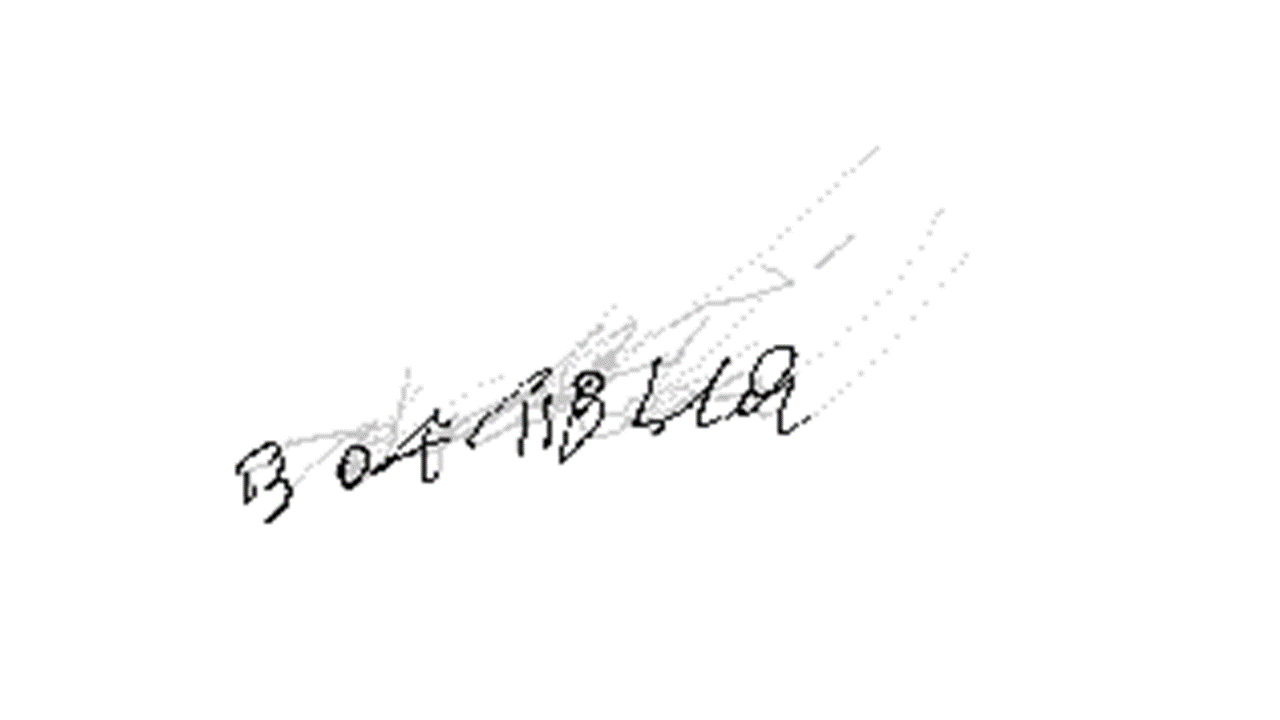} &
    \includegraphics[width=0.5\textwidth, height=4cm]{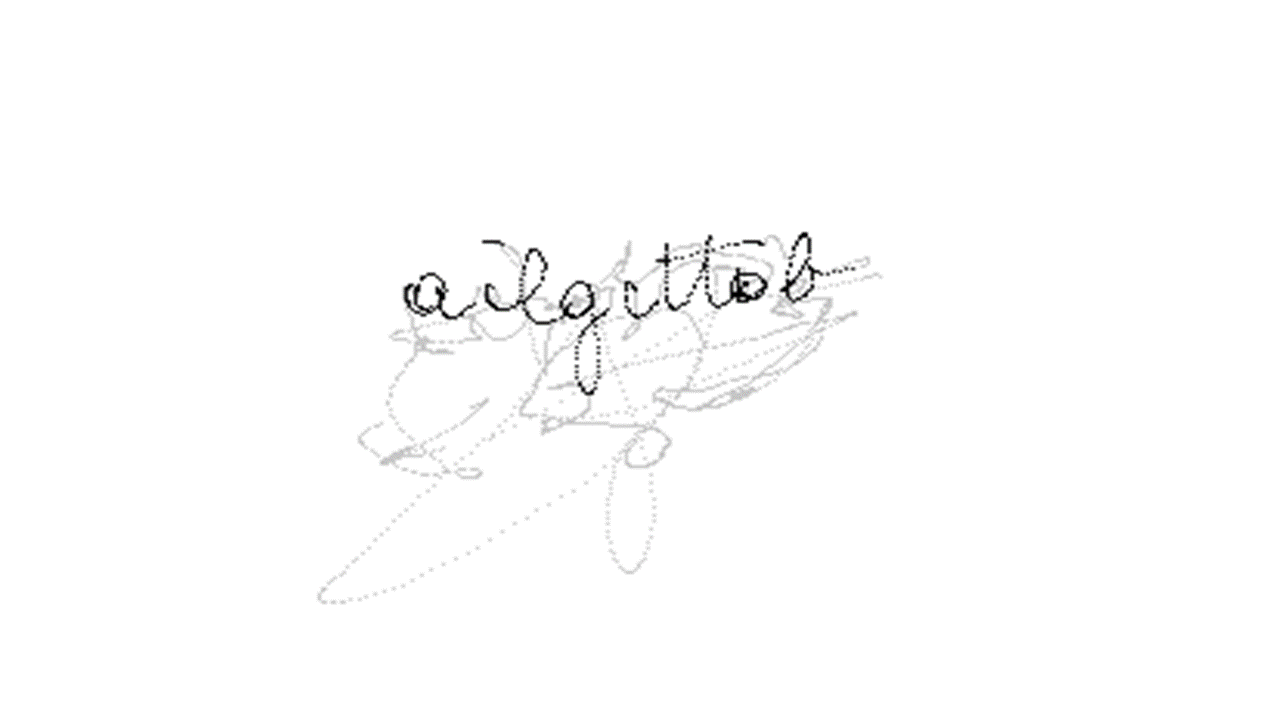} \\
  a) & b)  \\
  \end{tabular}\\
  \caption{On-paper (black) and in-air movements (gray) for the word "bottiglia" written in reverse by two different patients . a) AD patient 1; b) AD patient 2 }
    \label{fig:bottle_patient}
\end{figure}

First of all, contrary to what we expected, the model performed better without synthetic data at all. This surprising result could be explained by the extreme variability of in-air movements, which could not be modeled in the right way by the network used to generate synthetic data. The absence of visual feedback during in-air movements results in the patients' inability to control their movements and the generation of complex, almost random, in-air trajectories. Hence, predicting the probability distribution of the next in-air point is challenging. This is clear even visually, in Task 16, where the cognitive deficit resulted in very different forms of in-air movements (as many patients had difficulties in terms of motor programming when asked to write in reverse, see Figure~\ref{fig:bottle_patient}).
 
Next, the results show that the performance gap varies depending on the task. Task 16 involves a greater cognitive effort (writing backwards) than Task 13 and that results in the generation of longer and more complex in-air movements. On one hand, the greater cognitive effort of Task 16 makes the handwriting of AD patients more easily recognisable than the healthy controls’ handwriting when compared to the other task. On the other hand, the complexity of in-air movements has the drawback that a greater number of synthetic samples is required before they become beneficial with respect to the on-paper movements.


\section{Conclusion}~\label{conclusion}
In this work, we have explored the use of in-air movements for synthetic sample generation, particularly for a neurodegenerative disease like Alzheimer's disease. In accordance with the work in~\cite{in-air}, which states that in-air movements hold discriminative patterns, we have observed that indeed in-air movements have an impact in terms of model performance. 

We have observed that in-air movement quality and quantity depend on the nature of the task and the subject's motor and cognitive abilities, thus a subject/task-centered approach could lead to interesting results. Finally, further synthetic sample experiments could be done in the future to assess the model's performance with and without synthetic data. In addition, as future work, we plan to explore other methods for data generation, which may be more suitable for this particular task.

In summary, this work highlights the importance of exploring domain and data knowledge for improving data generation for health applications.


\section*{Acknowledgment}

This work has been partially supported by the Spanish project PID2021-126808OB-I00 (GRAIL) and the FI fellowship AGAUR 2020 FI-SDUR 00497 (with the support of the Secretaria d’Universitats i Recerca of the Generalitat de Catalunya and the Fons Social Europeu). The authors acknowledge the support of the Generalitat de Catalunya CERCA Program to CVC’s general activities.

\bibliographystyle{unsrt}
\bibliography{samplepaper.bib}
\end{document}